\documentclass{article}

\usepackage[dblblindworkshop, final]{neurips_2025}
\workshoptitle{Foundation Models for the Brain and Body}

\usepackage[utf8]{inputenc} 
\usepackage[T1]{fontenc}    
\usepackage{hyperref}       
\usepackage{url}            
\usepackage{booktabs}       
\usepackage{amsfonts}       
\usepackage{nicefrac}       
\usepackage{microtype}      
\usepackage{xcolor}         
\usepackage{amsmath}
\usepackage{graphicx}
\usepackage{subcaption}
\usepackage{caption} 
\usepackage{multirow}
\title{Learning the relative composition of EEG signals using pairwise relative shift pretraining}

\author{%
Christopher M. Sandino$^{1}$ \quad Sayeri Lala$^{1,2}$\thanks{Work completed during an internship at Apple.} \quad Geeling Chau$^{1,3}$\footnotemark[1] \quad Melika Ayoughi$^{1,4}$\footnotemark[1] \\ 
\textbf{Behrooz Mahasseni}$^1$ \quad \textbf{Ellen Zippi}$^1$ \quad \textbf{Ali Moin}$^1$ \quad \textbf{Erdrin Azemi}$^1$ \quad \textbf{Hanlin Goh}$^1$\\
$^1$Apple \quad $^2$Stanford University \quad $^3$California Institute of Technology \quad $^4$University of Amsterdam\\
\texttt{\{csandino,bmahasseni,ezippi,amoin,erdrin,hanlin\}@apple.com}\\
\texttt{slala21@stanford.edu} \quad \texttt{gchau@caltech.edu} \quad \texttt{m.ayoughi@uva.nl}
}

\begin{document}

\maketitle

\begin{abstract}
Self-supervised learning (SSL) offers a promising approach for learning electroencephalography (EEG) representations from unlabeled data, reducing the need for expensive annotations for clinical applications like sleep staging and seizure detection. While current EEG SSL methods predominantly use masked reconstruction strategies like masked autoencoders (MAE) that capture local temporal patterns, position prediction pretraining remains underexplored despite its potential to learn long-range dependencies in neural signals. We introduce \textbf{PA}irwise \textbf{R}elative \textbf{S}hift or \textbf{PARS} pretraining, a novel pretext task that predicts relative temporal shifts between randomly sampled EEG window pairs. Unlike reconstruction-based methods that focus on local pattern recovery, PARS encourages encoders to capture relative temporal composition and long-range dependencies inherent in neural signals. Through comprehensive evaluation on various EEG decoding tasks, we demonstrate that PARS-pretrained transformers consistently outperform existing pretraining strategies in label-efficient and transfer learning settings, establishing a new paradigm for self-supervised EEG representation learning.
\end{abstract}

\section{Introduction}\label{section:introduction}

Electroencephalography (EEG) signals provide non-invasive measurements of brain electrical activity that can reveal insights about sleep or neurological anomalies. While supervised learning approaches have shown expert-level performance in automatically decoding EEG for applications like sleep staging (\cite{Biswal2018}) and seizure detection (\cite{Jing2020}), they require large quantities of labeled data to train generalizable models. Self-supervised learning (SSL) presents a promising avenue to learn general representations from unlabeled EEG datasets, where encoder models are pretrained on pretext tasks with labels derived from the input signal itself. Previous studies have demonstrated that SSL pretraining reduces the need for labeled data during downstream task adaptation compared to traditional supervised methods (\cite{Cheng2020}, \cite{Banville2021}, \cite{Kostas2021}).

While SSL has been extensively studied in vision and natural language processing, optimal pretext tasks for EEG signals remain underexplored. Current approaches predominantly rely on masked reconstruction strategies, such as masked autoencoders (MAE), which learn representations by reconstructing masked temporal segments (\cite{He2022}, \cite{Chien2022}, \cite{Wang2024}, \cite{Wang2025}). Though effective, MAE-based methods primarily capture local temporal patterns necessary for reconstruction tasks. In contrast, position prediction pretraining—where models estimate the absolute or relative positions of segments within sequences—remains less explored in EEG literature despite its potential to learn long-range dependencies and capture the underlying temporal composition of neural signals (\cite{Zhai2022}, \cite{Wang2023}, \cite{Lala2024}, \cite{Ayoughi2025}).

\begin{figure}
  \centering
  \includegraphics[width=0.9\linewidth]{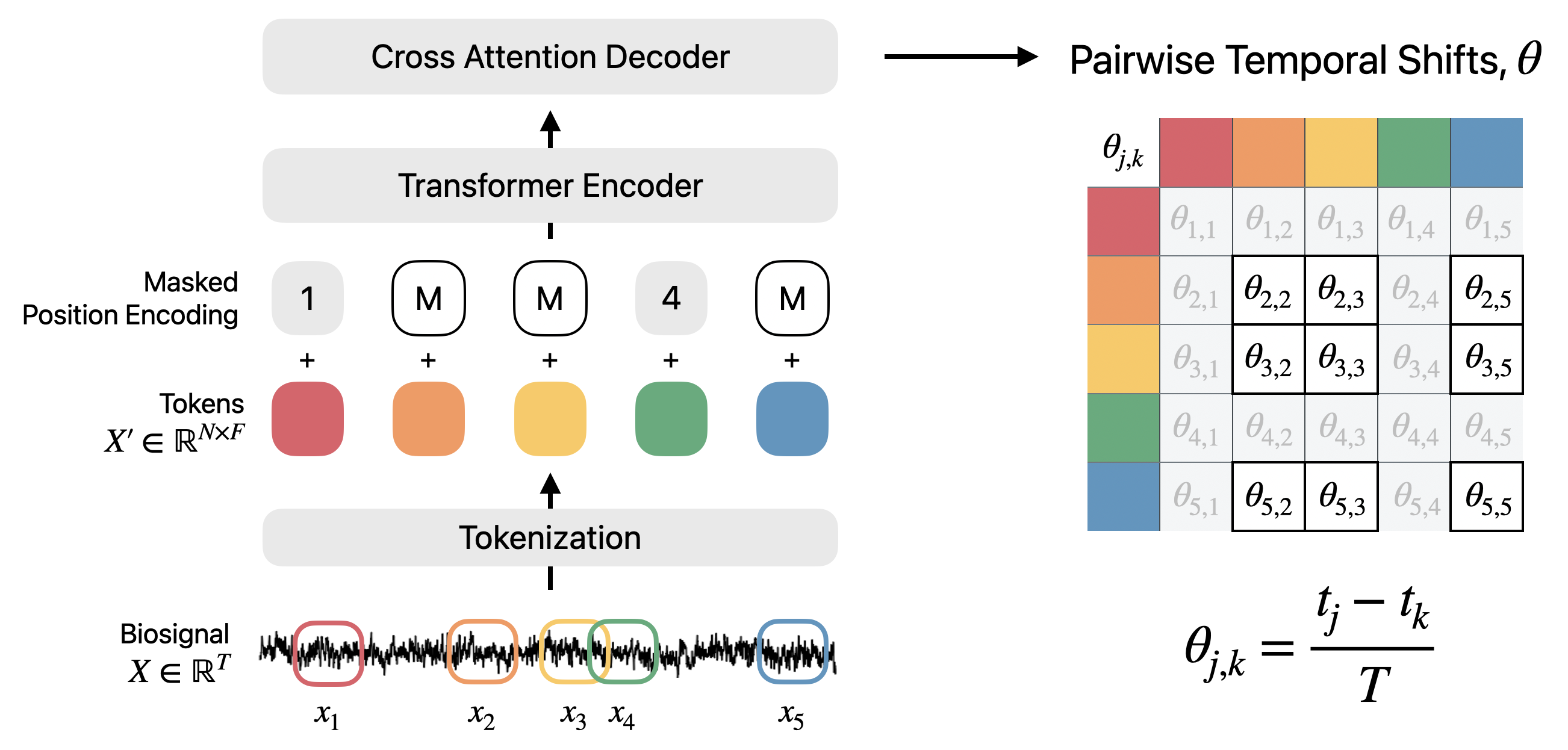}
  \caption{\textbf{PARS Pretraining Overview.} In PARS pretraining, $N$ patches are randomly sampled from a single-channel EEG sequence ($N=5$ is shown here for simplicity). Each patch is tokenized by a linear layer, and then positional embeddings are added to a subset of tokens. A learnable positional mask token (denoted by $M$) is added to the rest of the tokens. Tokens are embedded using a transformer encoder and then decoded by a cross-attention layer to estimate the distance matrix $\theta$ containing the temporal shifts between pairs of patches.}
  \label{fig:wise}
\end{figure}

In this study, we systematically compare pretraining strategies for EEG representation learning and introduce \textbf{PA}irwise \textbf{R}elative \textbf{S}hift or \textbf{PARS} pretraining, a novel pretext task inspired by the vision-based PART method (\cite{Ayoughi2025}) that predicts relative temporal shifts between randomly sampled EEG window pairs (as shown in Figure \ref{fig:wise}). Unlike reconstruction-based methods that focus on local pattern recovery, PARS encourages encoders to learn representations that capture relative temporal composition and long-range dependencies inherent in neural signals. Through comprehensive evaluation of various EEG decoding tasks, we demonstrate that PARS-pretrained transformers consistently outperform existing pretraining strategies in both label-efficient and transfer learning settings, establishing a new paradigm for self-supervised EEG representation learning.

\section{Pairwise Relative Shift (PARS)} 


\subsection{Tokenization and masked positional embedding}
Given a univariate sequence $X \in \mathbb{R}^{T}$ where $T$ is the number of samples, we randomly sample $N$ patches of fixed size $M$ to produce a series of patches $\{x_i\}_{i=1}^N$ where the $i$-th patch is denoted as $x_i \in \mathbb{R}^{M}$. Each $x_i$ is associated with a timestamp $t_i$ representing the temporal location of that patch in the original sequence $X$. 

Each patch in the series is tokenized by a learnable linear projection layer to produce a series of tokens $\{x_i'\}_{i=1}^N$ where $x_i' \in {\mathbb{R}^F}$ corresponds to the $i$-th token with dimensionality $F$. We adopt masked positional embedding (PE) from DropPos (\cite{Wang2023}), where a learnable position mask token is added to $N_m$ tokens, and the standard sinusoidal PE is added to the remaining $(N-N_m)$ tokens. The position masking ratio, $\gamma_{\text{pos}}=\frac{N_m}{N}$, is a hyperparameter that must be empirically tuned. 

\subsection{Pairwise relative shift estimation}
The pretext task in PARS is to estimate an anti-symmetric matrix $\theta \in \mathbb{R}^{N_m \times N_m}$ where each element $\theta_{j, k}$ corresponds to the relative normalized time shift between the $j$-th and $k$-th patch:
\begin{equation}\label{eqn:shifts}
    \theta_{j,k} = \frac{t_j - t_k}{T_s},
\end{equation}
Where $j$, $k$ correspond to indices of patches with masked PE, and $T_s$ is the total sequence length. The remaining pairs of patches without PE masking are excluded to avoid the leak of positional information that would make the pretext task too simple to learn meaningful representations. 

\subsection{Decoding with cross-attention}
A transformer encoder produces a series of embeddings $Y = \{y_i\}_{i=1}^N$ where $y_i$ is the embedding corresponding to the $i$-th patch. The decoder's task is to predict $\theta$ given a series of pairwise patch embeddings $Y' = \{[y_j, y_k]\}_{j, k}$ which are constructed by concatenating $y_j$ and $y_k$ along the feature dimension. A cross-attention decoder produces a series of embeddings, which are mapped by a learnable linear layer into the relative time shifts $\hat{\theta}$ using the following expression:
\begin{equation}\label{eqn:cross_attention}
\hat{\theta} = \texttt{cross\_attention}(Q=A_Q Y',K=Y,V=Y),
\end{equation}
where $Y$ is the series of patch embeddings, $Y'$ is the pairwise patch embeddings, and $A_Q \in \mathbb{R}^{2F \times F}$ is a linear projection layer that projects the pairwise patch embeddings into the same space as the patch embeddings. The cross-attention decoder shares information from all sampled patches amongst the pairwise embeddings being evaluated.

Finally, the transformer encoder and cross-attention decoder are trained end-to-end using a mean-squared error (MSE) loss between the decoder output and the relative normalized shift matrix:
\begin{equation}
L = ||\theta - \hat{\theta}||_2.
\end{equation}
Example reconstructions of the relative normalized shift matrix are shown in Figure \ref{fig:targets}.

\begin{figure}
  \centering
  \includegraphics[width=1.0\linewidth]{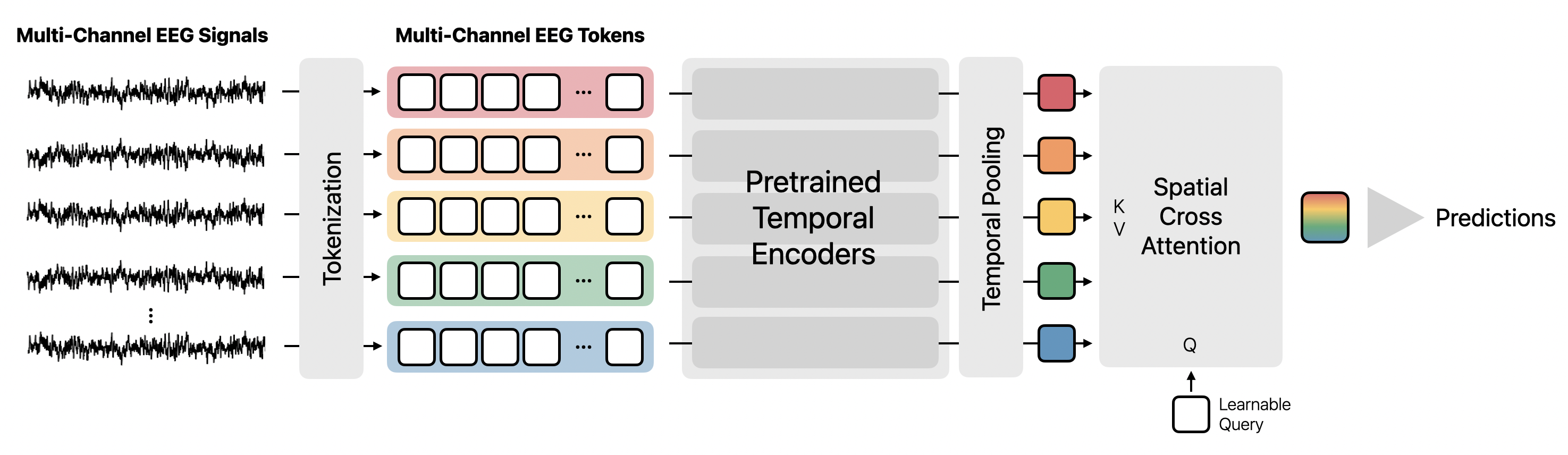}
  \caption{\textbf{Multi-channel Fine-tuning.} To adapt the PARS pretrained model for multi-channel EEG data, each channel of EEG is embedded using the tokenizer and single-channel transformer encoder learned during pretraining. The per-channel embeddings are average pooled along the temporal dimension to generate a series of spatial tokens. The spatial tokens are collapsed into a final token by applying a cross attention layer between them and a learnable query token. That final token is then converted into predictions by a linear layer.}
  \label{fig:mc_biot}
\end{figure}

\subsection{Multi-channel fine-tuning and evaluation}
The PARS pretraining approach produces an encoder for single-channel EEG data. To adapt the single-channel encoder for multi-channel EEG decoding tasks, we develop a new architecture inspired by prior works (\cite{Liu2022b}, \cite{Chau2025}) that embeds each channel using the pretrained single-channel encoder. Spatial information is then aggregated by a cross attention layer between the spatial tokens and a learnable query token. The final embedding is projected into predictions by a linear head. A diagram describing this multi-channel architecture is shown in Figure \ref{fig:mc_biot}.

\subsection{Implementation details}\label{section:implementation}

\begin{figure}
  \centering
  \includegraphics[width=1.0\linewidth]{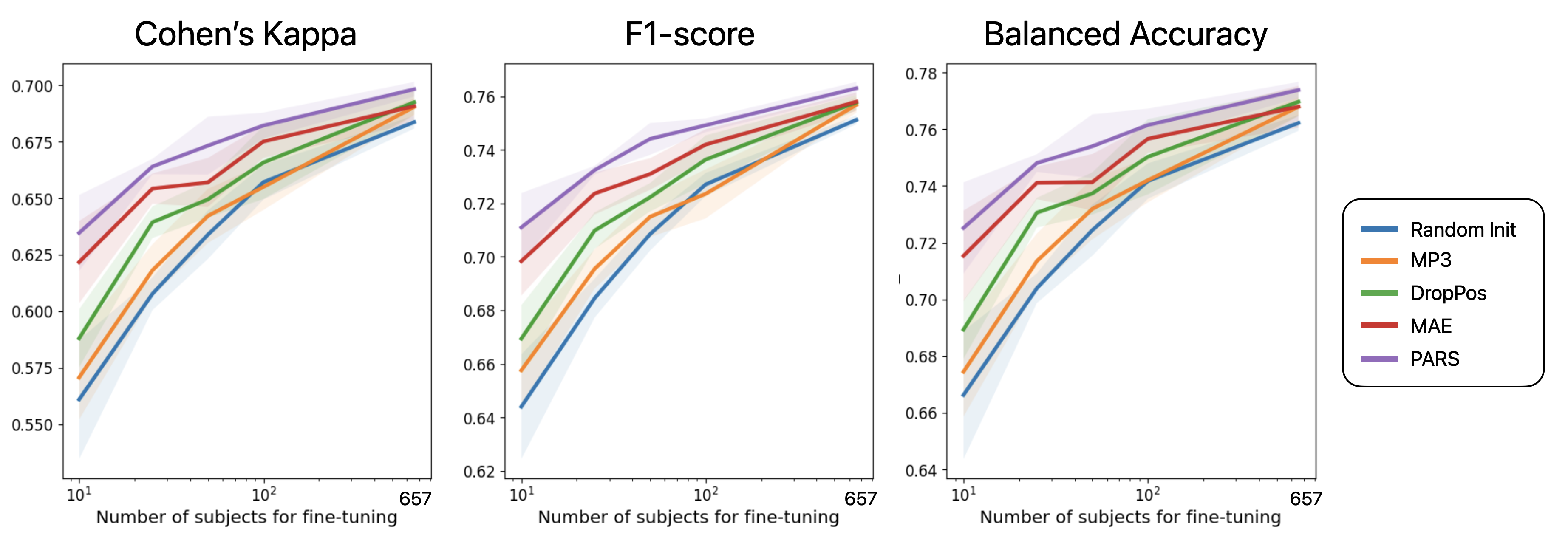}
  \caption{\textbf{Label Efficiency Experiment.} Clinical sleep staging (YSYW) results are shown for no pretraining (random init) and with pretraining using MP3, DropPos, MAE, and PARS. Each pretrained model is fine-tuned on a downsampled number of subjects depicted on the x-axis. Cohen's Kappa, F1-score, and balanced accuracy are reported on a test set of held out patients. The standard deviation across five random seeds is reported behind line plots for each approach.}
  \label{fig:results}
\end{figure}

\paragraph{Datasets}
For pretraining, we combine two large-scale clinical EEG databases: The You Snooze You Win Challenge Dataset (\cite{Ghassemi2018}), and the Temple University Hospital EEG Corpus (\cite{Obeid2016}) to form a dataset containing over 1.5 million unique 30-second examples. The data pre-processing used to prepare each database for pretraining is detailed in Appendix \ref{appendix:pt_datasets}.

\paragraph{Model Architecture}
For the temporal encoder architecture, we use a transformer model with 12.7M parameters for all pretraining algorithms evaluated in this work. The encoder contains 8 transformer blocks comprised of multi-head attention blocks (\cite{Vaswani2017}), each with 8 heads, and feed-forward blocks with a hidden dimension of size 512. A more detailed description of the encoder architecture used in this work is provided in Appendix \ref{appendix:biot}.

\paragraph{Pretraining}
During pretraining, small 1-sec patches are randomly sampled from 30-sec long EEG signals, and converted into tokens of size 512. Through a hyperparameter search described in Appendix \ref{appendix:ablations}, we find that the optimal position masking ratio $\gamma_{\text{pos}}$ is 0.8, and the number of patches $N$ is 40. PARS pretraining is performed for 1000 epochs using the AdamW optimizer (\cite{Loshchilov2019}) with a learning rate of $1.0 \times 10^{-4}$, weight decay of $1.0 \times 10^{-4}$, and batch size of 512. We use a learning rate scheduler with linear warmup from 10\% of the maximum learning rate for the first 100 epochs, followed by cosine annealing for the remaining 900 epochs. The source signals are augmented by random cropping and randomly selecting one channel from multi-channel recordings in the pretraining dataset.

\paragraph{Fine-tuning}
During fine-tuning, the cross attention decoder is replaced with learnable spatial cross attention and linear layers as depicted in Figure \ref{fig:mc_biot}. All model layers are fully fine-tuned using a weighted cross-entropy loss where the weights are adjusted by class frequency. Unlike in the PARS pretraining phase, patches are sampled using a fixed sampling with a 1-sec stride, and masked sinusoidal PE is replaced with a standard sinusoidal PE. All fine-tuning experiments are performed for 200 epochs using the AdamW optimizer with the same learning rate and weight decay settings as pretraining. The datasets used for fine-tuning are relatively small, making the model susceptible to overfitting during fine-tuning stages. To mitigate overfitting, spatial tokens are randomly dropped with 0.5 probability before the spatial cross attention layer. The model checkpoint with the lowest validation loss is chosen to evaluate on the test set. 

\paragraph{Compute Resources}\label{section:compute}
Both pretraining and fine-tuning experiments are performed on nodes with 40 cores and four NVIDIA Tesla V100 32GB cards. All code is implemented using PyTorch (\cite{Paszke2019}). All evaluation metrics are computed using the TorchMetrics library (\cite{Detlefsen2022}).


\section{Experiments \& Results}

We evaluate the PARS pretrained model in two experimental settings: 1) label-efficient learning and 2) transfer learning. In both experiments, we compare PARS against four other approaches: 1) Supervised learning without pretraining, 2) MAE: Masked Autoencoder (\cite{He2022}), 3) MP3: Masked Position Prediction (\cite{Zhai2022}), and 4) DropPos: Dropped Position Prediction (\cite{Wang2023}). The details of these pretraining methods are described in Appendix \ref{appendix:baselines}. 

\paragraph{Label Efficiency} Each pretrained model is used to warm start a fine-tuning experiment using the YSYW dataset which is further described in Appendix \ref{appendix:ft_datasets}. To evaluate the label efficiency of each approach, we vary the number of subjects in the training set during fine-tuning from 10 to 657. Each fine-tuning experiment is repeated five times with different random seeds, and a different set of random patients from the training set. For each seed and each pretrained model, we evaluate performance on a test set of 117 held out subjects, and report the average and standard deviation of three decoding accuracy metrics in Figure \ref{fig:results}.

The four pretraining approaches yield different label efficiency profiles. All pretraining algorithms significantly outperform the supervised learning approach without pretraining. In the very low label regime (10 subjects), PARS outperforms the other three pretraining strategies indicating that relative temporal position information is more important than absolute position information for the sleep staging task. As the number of labels increases, disparity between pretraining algorithms decreases as sleep staging accuracy becomes less reliant on the initialization from pretraining.

\begin{table}[]
\centering
\setlength{\tabcolsep}{3pt}
\scriptsize
\tiny
\begin{tabular}{lcccccccccccc}
\toprule
         && \multicolumn{2}{c}{\bf EESM17, 5-class} && \multicolumn{2}{c}{\bf TUAB, 2-class} && \multicolumn{2}{c}{\bf TUSZ, 2-class} && \multicolumn{2}{c}{\bf PhysioNet-MI, 4-class} \\ \cline{3-4} \cline{6-7}  \cline{9-10} \cline{12-13}
         && Bal. Acc.      & Kappa     && Bal. Acc.       & AUROC       && Bal. Acc.            & AUROC            && Bal. Acc.           & Kappa            \\ \midrule
\multicolumn{12}{l}{\bf Supervised-only baseline}\\[0.5em]
Scratch  && 0.733 $\pm$ 0.092 & 0.637 $\pm$ 0.111 && 0.782 $\pm$ 0.002 & 0.868 $\pm$ 0.002  && 0.834 $\pm$ 0.018   & 0.840 $\pm$ 0.009    && 0.527 $\pm$ 0.008       & 0.369 $\pm$  0.011       \\\midrule
\multicolumn{12}{l}{\bf Reconstruction-based}\\[0.5em]
MAE      && \underline{0.756 $\pm$ 0.131} & \underline{0.672 $\pm$ 0.156} && \underline{0.801 $\pm$ 0.004} & 0.872 $\pm$ 0.004  && \underline{0.896 $\pm$ 0.012} & \bf 0.894 $\pm$ 0.009  && \underline{0.540 $\pm$  0.043}      & \underline{0.386 $\pm$ 0.057}      \\\midrule
\multicolumn{12}{l}{\bf Position-prediction-based}\\[0.5em]
MP3      && 0.751 $\pm$ 0.104 & 0.659 $\pm$ 0.129 && 0.792 $\pm$ 0.003 & 0.872 $\pm$ 0.003  && 0.863 $\pm$ 0.009   & 0.847 $\pm$ 0.005    && 0.486 $\pm$ 0.078      & 0.314 $\pm$  0.104      \\
DropPos  && 0.752 $\pm$ 0.129 & 0.662 $\pm$ 0.159 && 0.795 $\pm$ 0.002 & \underline{0.875 $\pm$ 0.002} && 0.872 $\pm$ 0.007  & 0.840 $\pm$ 0.009    &&  0.533 $\pm$  0.009      & 0.378 $\pm$ 0.012       \\
PARS     && \bf 0.758 $\pm$ 0.130 & \bf 0.674 $\pm$ 0.154 && \bf 0.802 $\pm$ 0.003 & \bf 0.877 $\pm$ 0.003  && \bf 0.901 $\pm$ 0.009 & \underline{0.878 $\pm$ 0.007}  && \bf 0.572 $\pm$ 0.007      & \bf  0.429 $\pm$ 0.009       \\\bottomrule
\end{tabular}
\caption{\textbf{Transfer Learning Results.} Metrics are reported over each dataset's respective test set, which is held out from pretraining, fine-tuning, and validation stages. Best is bolded and second best is underlined.}
\label{tab:results}
\end{table}


\paragraph{Transfer Learning}\label{section:transfer_learning}

We evaluate the adaptability of each pretrained model by transfer learning to four different datasets: 1) Wearable Sleep Staging (EESM17), 2) Abnormal EEG Detection (TUAB), 3) Seizure Detection (TUSZ), and 4) Motor Imagery (PhysioNet-MI). The details of each dataset and pre-processing are provided in Appendix \ref{appendix:ft_datasets}. We selected these four datasets to test model performance across diverse EEG acquisition parameters such as number of channels, referencing schemes, window lengths, and electrode placement (e.g. scalp and in-ear). For TUAB, TUSZ, and PhysioNet-MI, the fine-tuning and evaluation process is repeated five times with different seeds to calculate the average and standard deviation for each metric. EESM17 is a relatively smaller dataset, so we instead follow the leave one subject out cross-validation process used by \cite{Mikkelsen2017}. In this process, average and standard deviation of performance metrics are calculated across subjects instead of seeds. Per-subject metrics are reported in Figure \ref{fig:eesm_results}.

As shown in Table \ref{tab:results}, all pretraining strategies demonstrate impressive generalizability across the four datasets. The PARS pretrained model outperforms the other three pretraining models on three out of four tasks. However, MAE and PARS perform similarly for the TUSZ task. This suggests that the local features being learned by MAE may be slightly better for the seizure detection task compared to the global features being learned by PARS. Prior work suggests that features learned by pretraining strategies such as MAE and position prediction can be complementary (\cite{Zhai2022}). Further investigation is required to determine whether these pretext tasks can be combined to learn even more powerful representations that can outperform MAE and PARS alone. 

\section{Conclusion}\label{section:conclusion}

Our proposed PARS pretraining method successfully captures long-range temporal dependencies by learning relative positional relationships between EEG patches, leading to superior performance across multiple decoding tasks compared to existing SSL strategies. The consistent improvements observed across tasks suggest that temporal composition awareness is a crucial inductive bias for EEG analysis. These findings open new research directions for self-supervised learning in neuroscience applications, where understanding temporal structure is fundamental to decoding brain activity. Future work could explore hybrid approaches that combine the local pattern sensitivity of reconstruction methods with the global temporal awareness of position prediction, potentially unlocking even richer representations for advancing automated EEG analysis in clinical and research settings.

\bibliographystyle{plainnat}
\bibliography{neurips_2025}

\newpage
\appendix
\section{Appendix}

\subsection{Encoder Architecture}\label{appendix:biot}

\begin{figure}[h]
  \centering
  \includegraphics[width=0.5\linewidth]{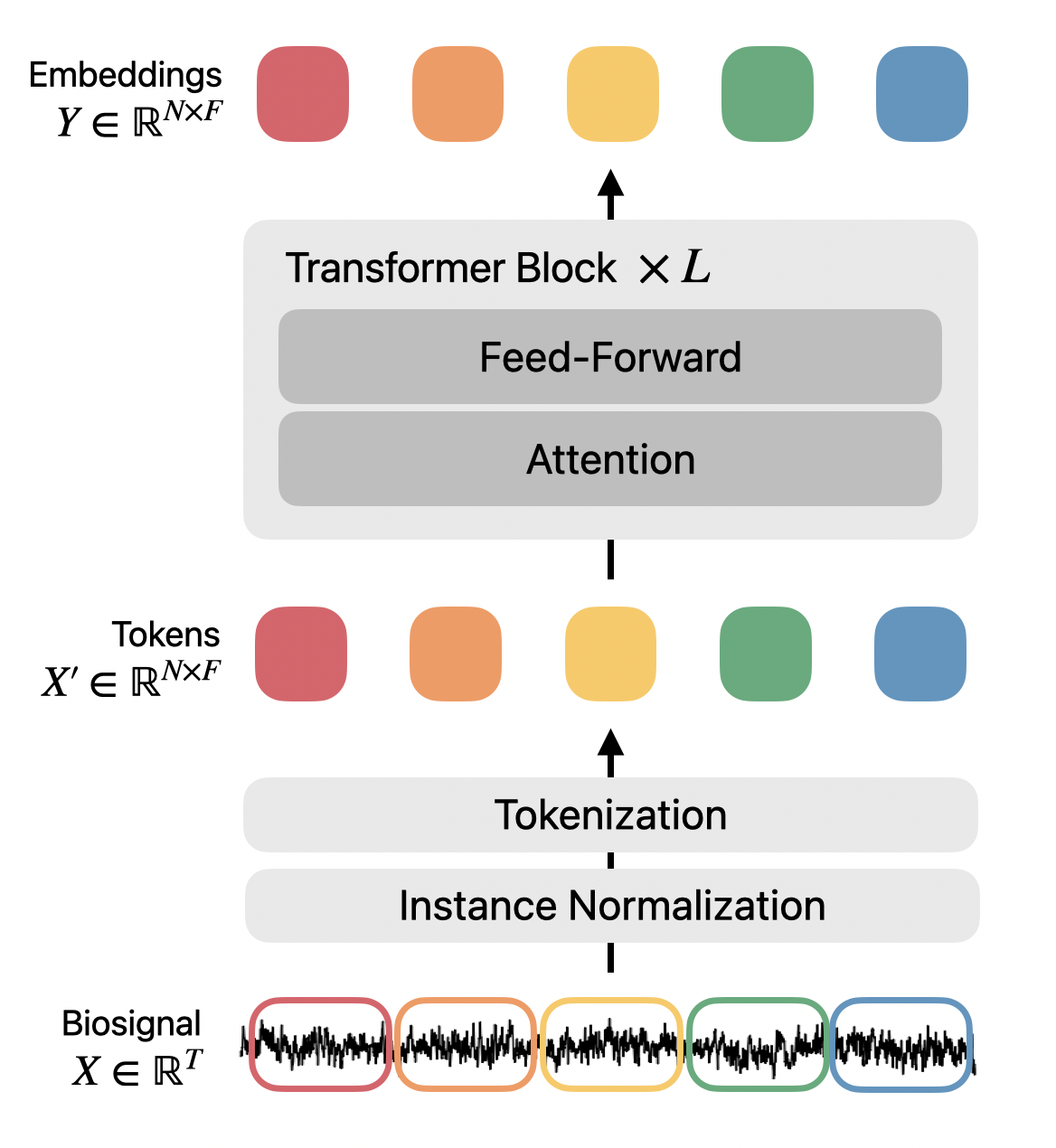}
  \caption{\textbf{Transformer encoder architecture.} This figure depicts the transformer encoder model architecture used for all pretraining experiments in the paper.}
  \label{fig:biot}
\end{figure}

The main model architecture used in this work is a transformer encoder based on PatchTST presented by \cite{Nie2023}. The input sequence is instance normalized, divided into $N$ patches, and tokenized by a linear projection layer into tokens of size 512. During fine-tuning, patches are always sampled uniformly in a non-overlapping fashion. The weights of the linear tokenizer are learned during the pretraining process. Tokens are processed by transformer blocks that implement the following update rule:
\begin{align}
x'' &= x' + \text{MHSA}(\text{LN}(x')) \\
y &= x'' + \text{FF}(\text{LN}(x'')
\end{align}
where tokens are processed sequentially by layer normalization (LN), multi-head self-attention blocks (MHSA), and feed-forward (FF) blocks. Feed-forward blocks consist of two linear layers with a GeLU activation in between (\cite{Hendrycks2016}). A diagram of the model architecture is shown in Figure \ref{fig:biot}.


\subsection{Baseline Approaches}\label{appendix:baselines}
\begin{figure}[h]
  \centering
  \includegraphics[width=1.0\linewidth]{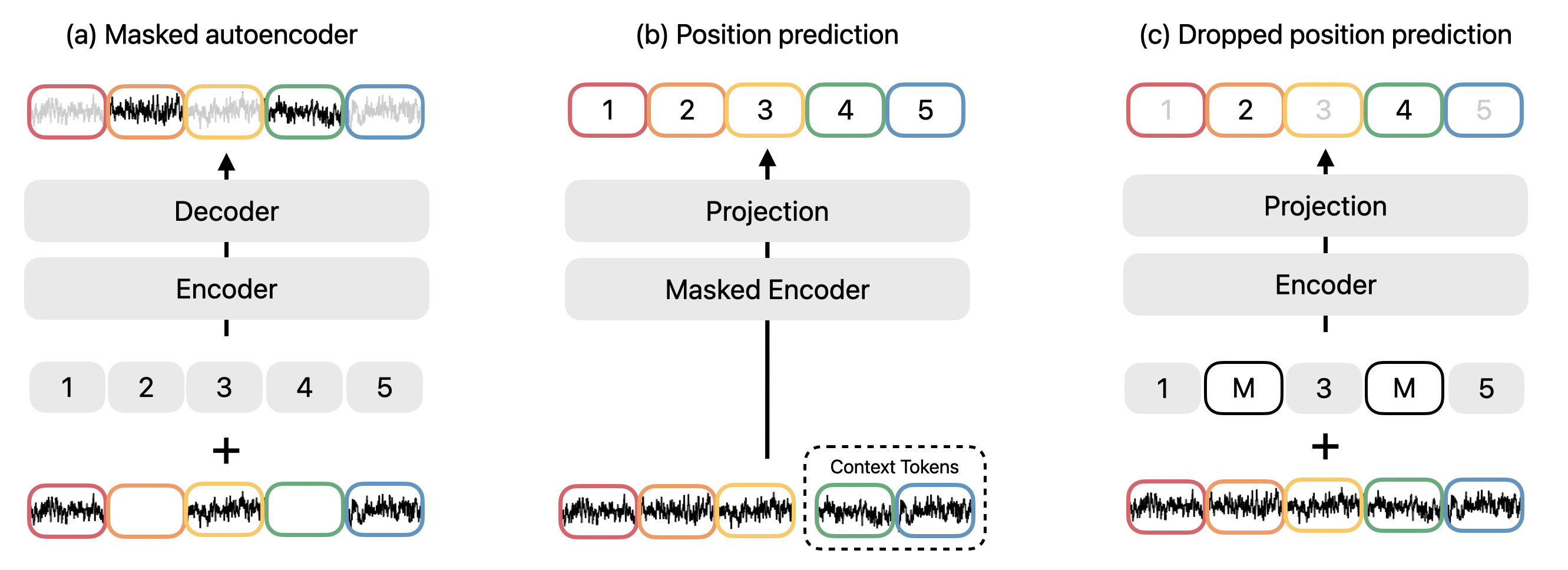}
  \caption{A comparison of pretraining strategies.}
  \label{fig:ssl}
\end{figure}

Here, we explain the implementation details of baseline pretraining approaches that we compared against PARS pretraining. The hyperparameters, such as masking ratio and position drop rate, are tuned for each approach using the YSYW validation set. A diagram of each pretraining approach is shown in Figure \ref{fig:ssl}.

\paragraph{Masked Autoencoder} The Masked Autoencoder (MAE) pretraining approach is adapted from the implementation described in \cite{He2022}. The input sequence is patchified and then tokenized into $N$ tokens, randomly masked with masking ratio $\gamma$. Before masking, a fixed sinusoidal position embedding is added so the encoder has information about each token's temporal position. Masked tokens are embedded using an 8-layer transformer encoder. The embeddings are rearranged into a sequence of size $N$ containing the embeddings in their original positions and learnable mask tokens in the masked tokens' locations. Finally, a single-layer transformer decoder reconstructs the rearranged embedding sequence into the original time domain. The entire MAE is pretrained by minimizing the element-wise mean squared error loss between the decoder output and the original input sequence.

\paragraph{MP3} The Masked Position Prediction Pretraining (MP3) approach is adapted from the implementations described in \cite{Zhai2022}. The input sequence is patchified and tokenized into $N$ tokens, which are shuffled to break the temporal ordering of the tokens. An 8-layer transformer encoder with a single linear classification head is used to classify each token into its original absolute position in the sequence without using positional embeddings. The MP3 pretext task is made more difficult by masking a subset of the input tokens from the keys and values of each attention layer. The encoder and projection head are trained using a cross-entropy loss with $N$ position classes.

\paragraph{DropPos} The Dropped Position (DropPos) prediction pretraining approach is adapted from the implementation described in \cite{Wang2023}. Like MAE, the input sequence is patchified and tokenized into $N$ tokens, randomly masked with masking ratio $\gamma$. A masked sinusoidal positional embedding is added to the remaining tokens consisting of fixed sinusoidal positional embedding when positions are kept and a learnable position mask token when positions are dropped. A separate ratio controls the position drop rate denoted as $\gamma_{\text{pos}}$. Finally, the DropPos pretext task goal is to reconstruct the dropped positions using an 8-layer transformer encoder with a linear classification head. The encoder and classification head are trained using a cross-entropy loss with $\gamma_{\text{pos}} (1 - \gamma) N$ position classes.

\subsection{Datasets}

\subsubsection{Pretraining}\label{appendix:pt_datasets}
For all pretraining experiments, we combine the following two databases into a single pretraining set:

\begin{itemize}
\item The You Snooze You Win Challenge Dataset (YSYW\footnote[1]{https://physionet.org/content/challenge-2018/1.0.0/}) is a large public polysomnography (PSG) database collected from 1,983 patients undergoing sleep studies at Massachusetts General Hospital (\cite{Ghassemi2018}, \cite{Goldberger2000}). Each PSG recording contains six EEG channels measured at the following 10-20 locations: F3, F4, C3, C4, O1, O2. All data is bandpass filtered (0.3 - 75 Hz, FIR design) and notch filtered to remove 60 Hz interference using SciPy (\cite{Virtanen2020}). The YSYW dataset contains two sets of patients: 994 with annotations, and 989 without. For the experiments in this paper, we combine the two sets of patients into the pretraining set, but hold out some patients with annotations for validation and testing as described in Appendix \ref{appendix:ft_datasets}. Ultimately, we use overnight recordings of 1,396 patients for pretraining. YSYW is under Open Data Commons Attribution License v1.0.

\item The Temple University Hospital EEG Corpus (TUEG\footnote[2]{https://isip.piconepress.com/projects/nedc/html/tuh\_eeg/\#c\_tueg}) is a large clinical database of EEG recordings from 14,987 patients (\cite{Obeid2016}). We take the common 19 EEG channels measured at the following 10-20 locations: Fp1, Fp2, F3, F4, F7, F8, FZ, C3, C4, CZ, P3, P4, PZ, O1, O2, T3, T4, T5, T6. All channels are resampled to 200 Hz from their original sampling rate, which varies across recordings. Then data is bandpass filtered (0.3 - 75 Hz), and notch filtered to remove 60 Hz interference. The fine-tuning datasets described in Appendix \ref{appendix:ft_datasets} such as TUAB and TUSZ are subsets of the much larger TUEG dataset, so we discard all TUAB/TUSZ patients from the pretraining. 

\end{itemize}

\subsubsection{Fine-tuning}\label{appendix:ft_datasets}

Below are details for each of the fine-tuning datasets used to evaluate PARS and other pretraining approaches in Table \ref{tab:results}.

\begin{table}
\centering
\begin{tabular}{lllll}
\toprule
Dataset & \# Subjects & \# Samples & Channels     & Task  \\ \midrule
YSYW     & 1,386       & 756,563   & 6   & Pretraining          \\
TUH      & 14,987      & 768,695   & 19  & Pretraining          \\ \midrule
YSYW     & 994         & 292,690   & 6   & Sleep Staging (Clinical)        \\
EESM17   & 9           & 7,390     & 12  & Sleep Staging (Wearable)       \\
TUAB     & 2,329       & 59,524    & 21  & Abnormal Detection    \\
TUSZ     & 675         & 38,610    & 19  & Seizure Detection    \\
PhysioNet-MI  & 109    & 9,837     & 64  & Motor Imagery Classification    \\

\bottomrule
\end{tabular}
\caption{Dataset statistics.}
\label{tab:datasets}
\end{table}

\begin{itemize}    
    \item \textbf{Clinical Sleep Staging}: For the fine-tuning experiments in this work, the 994 annotated patients from YSYW are randomly sub-divided into training, validation, and test sets using a 60/20/20 split. We focus our analysis on classifying 30-sec long sequences from six-channel scalp EEG into one of five stages: Wake, REM, Non-REM1, Non-REM2, and Non-REM3. Sequences labelled as ``undefined'' are discarded from fine-tuning and evaluation. Sleep staging accuracy is evaluated using balanced accuracy and Cohen's Kappa, a standard metric used to quantify sleep staging performance across raters (\cite{Lee2022}).
    
    \item \textbf{Wearable Sleep Staging}: The Ear-EEG Sleep Monitoring 2017 (EESM17\footnote[3]{https://openneuro.org/datasets/ds004348/versions/1.0.5}) dataset contains overnight recordings from 9 subjects with a 12-channel wearable ear-EEG system and a 6-channel scalp-EEG system (\cite{Mikkelsen2017}). The data contains sleep stage labels that were annotated by experts using the scalp-EEG data. We focus our analysis on classifying 30-sec long sequences from the 12-channel ear-EEG system into one of five stages: Wake, REM, Non-REM1, Non-REM2, and Non-REM3. Segments labelled as ``invalid'' are excluded from the analysis. All channels are bandpass filtered (0.3 - 75 Hz), and notch filtered to remove 50 Hz interference. To allow the model to better generalize to subjects with variable skin-electrode contact quality across channels, we also add Gaussian noise to randomly selected channels during fine-tuning. Sleep staging accuracy is evaluated across each of the folds using balanced accuracy and Cohen's Kappa. EESM17 is under Creative Commons CC0 license.

    \item \textbf{Abnormal Detection}: The Temple University Hospital Abnormal EEG Corpus v3.0.1 (TUAB\footnote[4]{https://isip.piconepress.com/projects/nedc/html/tuh\_eeg/\#c\_tuab}) is a subset of TUEG that contains recordings from 2329 unique patients each of which is labelled as ``normal'' or ``abnormal'' (\cite{Lopez2015}). We focus our analysis on performing abnormal detection from 30-sec long sequences measured from 21 common EEG channels (Fp1, Fp2, F3, F4, F7, F8, FZ, C3, C4, CZ, P3, P4, PZ, O1, O2, T3, T4, T5, T6, A1, A2). All channels are resampled to 200 Hz from their original sampling rate, which varies across recordings. All channels are bandpass filtered (0.3 - 75 Hz), and notch filtered to remove 60 Hz interference. Patients are divided into training and test sets using splits provided within the dataset. The training set is further sub-divided into training and validation splits using an 80/20 split. Abnormal detection accuracy is evaluated using accuracy and area under the receiver operating characteristic curve (AUROC).
    
    \item \textbf{Seizure Detection}: The Temple University Hospital EEG Seizure Corpus v2.0.3 (TUSZ\footnote[5]{https://isip.piconepress.com/projects/nedc/html/tuh\_eeg/\#c\_tusz}) is a subset of TUEG that contains recordings from 675 patients with start and end locations of seizure events (\cite{Shah2018}). We focus our analysis on performing binary seizure detection from 30-sec long sequences measured from 19 common EEG channels (Fp1, Fp2, F3, F4, F7, F8, FZ, C3, C4, CZ, P3, P4, PZ, O1, O2, T3, T4, T5, T6). All channels are resampled to 200 Hz from their original sampling rate, which varies across recordings. Additionally, all channels are bandpass filtered (0.3 - 75 Hz) and notch filtered to remove 60 Hz interference. Patients are divided into training and test sets using splits provided within the dataset. The training set is further sub-divided into training and validation splits using an 80/20 split. Seizure detection performance is evaluated using accuracy and AUROC.

    \item \textbf{Motor Imagery}: The PhysioNet Motor Imagery (PhysioNet-MI\footnote[6]{https://physionet.org/content/eegmmidb/1.0.0/}) is an EEG motor imagery dataset consisting of 109 subjects, each with recordings of either left fist, right fist, both fists, or both feet motor imagery. We focus our analysis on performing multi-class classification from 4-sec long sequences measured from 64 common EEG channels. All channels are resampled to 200 Hz from their original sampling rate of 160 Hz, high-pass filtered with 0.3 Hz, 60 Hz notch filtered, and averaged referenced. We use subjects 1-70 for training, 71-89 for validation, and 90-109 for test. Performance is evaluated using Cohen's Kappa and Accuracy. PhysioNet-MI is under Open Data Commons Attribution License v1.0.
    
\end{itemize}

\subsection{Hyperparameter Tuning \& Ablations}\label{appendix:ablations}

\begin{figure}[h]
  \centering
  \includegraphics[width=1.0\linewidth]{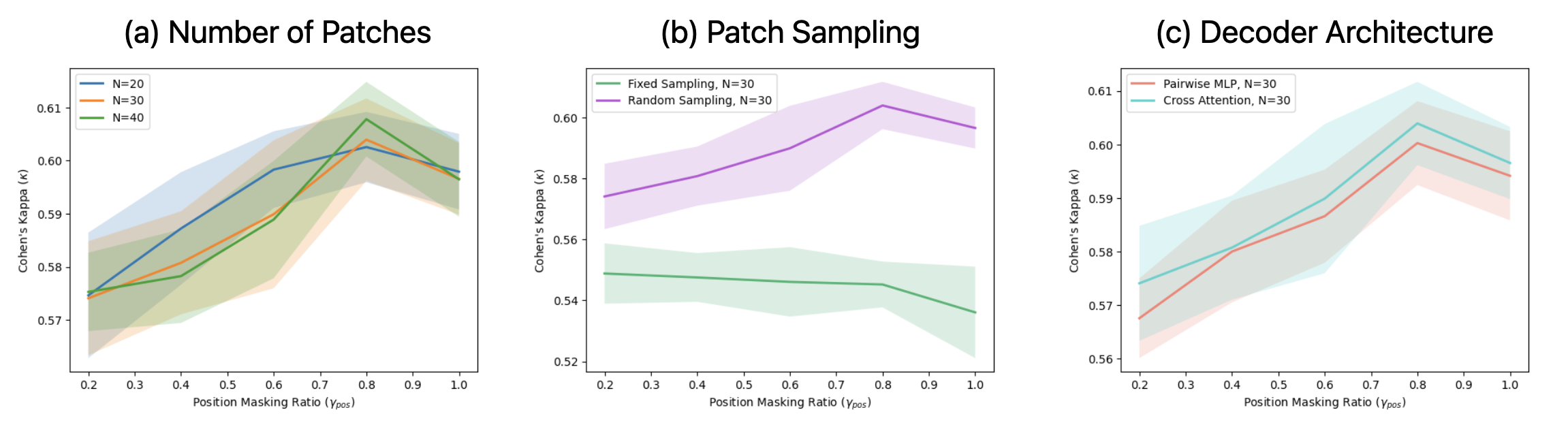}
  \caption{\textbf{Hyperparameter Tuning and Ablations.} (a) Downstream Cohen's Kappa is reported for varying values of the number of patches $N$ and position masking ratio $\gamma_{\text{pos}}$. (b) Similarly, we plot sleep staging accuracy for (b) fixed patch sampling vs. random sampling and (c) pairwise MLP vs. cross attention decoder architectures. The standard deviation of Cohen's Kappa is computed across five random seeds and plotted behind each line plot.}
  \label{fig:ablations}
\end{figure}

To study the effect of various hyperparameters and design choices in PARS, we provide results from ablation studies evaluated using the YSYW dataset. For all ablations, we pretrain a transformer encoder using the full YSYW and TUEG sets described in Section \ref{appendix:pt_datasets}. After pretraining, we fine-tune each model on a subset of the YSYW training set comprised of 100 subjects. This process is repeated five times with different seeds and different training subjects. Finally, each fine-tuned model is evaluated by computing the average Cohen's Kappa across 200 subjects from the validation set. Results across seeds are averaged and plotted in Figure \ref{fig:ablations}.

\paragraph{Masking} During pretraining, we vary the number of patches $N$ (20, 30, 40) and PE mask ratio $\gamma_{\text{pos}}$ (0.2, 0.4, 0.6, 0.8, 1.0). As demonstrated by the pretraining loss curves in Figure \ref{fig:ablations}a, the pretraining task becomes easier as the number of patches $N$ is increased. Intuitively this makes sense because when more patches are sampled, less of the signal is masked, thus making it easier for the cross-attention decoder to reconstruct pairwise temporal distances. The pretraining task difficulty is also increased as $\gamma_{\text{pos}}$ increases, ultimately leading to better downstream performance. However, it is important to note that as $N$ and $\gamma_{\text{pos}}$ increase linearly, the number of pairs seen by the decoder increases combinatorially and therefore limiting this approach by the amount of available GPU memory.

\paragraph{Patch Sampling} We explore the effectiveness of random sampling patches from the input EEG signal by comparing random sampling to fixed sampling with a 1-sec stride. As shown in Figure \ref{fig:ablations}b, random sampling performs significantly better than fixed sampling across a variety of $\gamma_{\text{pos}}$ values. The pretext task of estimating temporal shifts between randomly sampled patches is harder than for uniformly spaced patches due to the larger number of variations in inter-patch distances. Due to its increase in task difficulty, random sampling leads to learning of stronger representations for downstream decoding tasks.

\paragraph{Decoder Architecture} We explore two different decoder architectures to estimate the distance matrix $\theta$ given the pairwise patch embeddings. The first architecture is a pairwise multi-layer perceptron (MLP) that takes in a concatenated pairwise patch embedding $[y_j, y_k]$ and outputs the relative shift between the $j$-th and $k$-th patch. The MLP consists of two linear layers with hidden size of 512 and ReLU activation in between. The pairwise MLP is compared against the cross attention decoder as described by Eqn. \ref{eqn:cross_attention}. As shown in Figure \ref{fig:ablations}c, the cross attention decoder outperforms the pairwise MLP since it has access to all patches through the cross attention mechanism, whereas the pairwise MLP only has access to a single pair embedding instance.

\subsection{Experimental Results}\label{appendix:results}

\begin{figure}[h]
  \centering
  \includegraphics[width=0.9\linewidth]{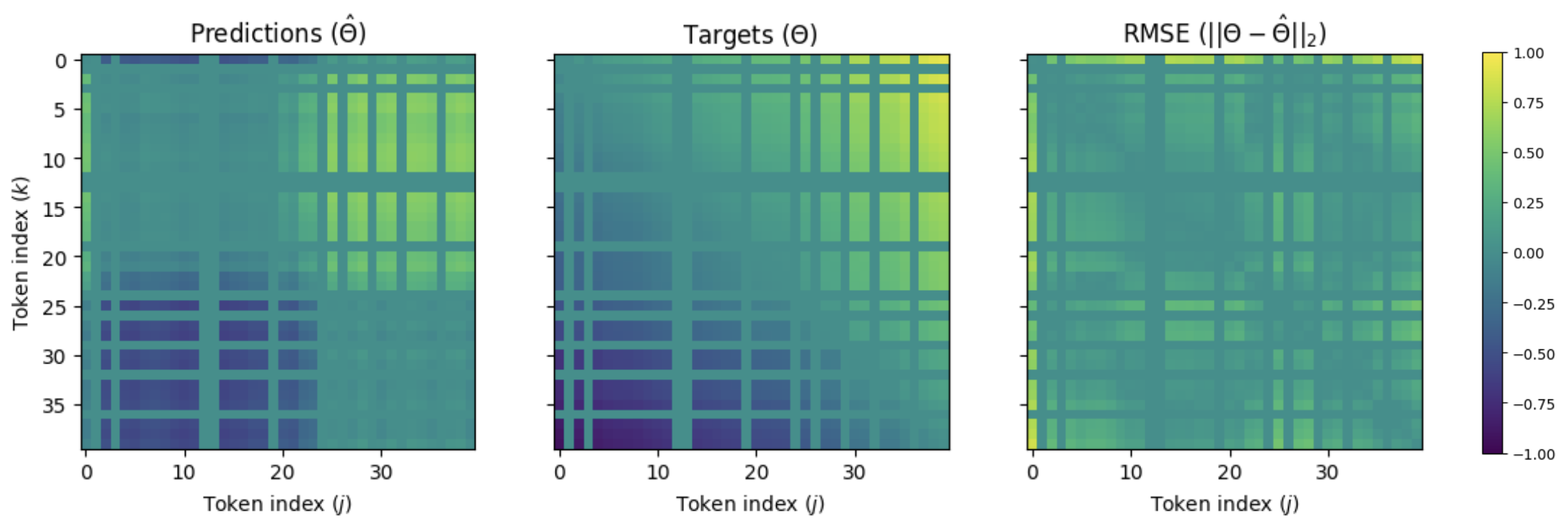}
  \caption{\textbf{Relative Time Shift Reconstructions.} Example reconstructions $\hat{\theta}$ (left) with corresponding targets $\theta$ (middle), and their root mean squared error (right). In this case, the position masking ratio is 80\%, so 32 $\times$ 32 pairwise distances must be reconstructed from the total 40 $\times$ 40 distances shown in $\theta$. Some rows and columns are zero because these correspond to patches with positional embedding. The distances for these pairs of patches are not considered in the loss. Targets are bounded by $[-1, 1]$ because they are normalized by the maximum signal extent, as Eqn. \ref{eqn:shifts} shows. Note that tokens are sorted in chronological order to better visualize predictions and targets, but in reality, the token order is always shuffled during training.}
  \label{fig:targets}
\end{figure}

\begin{figure}[h]
  \centering
  \includegraphics[width=1.0\linewidth]{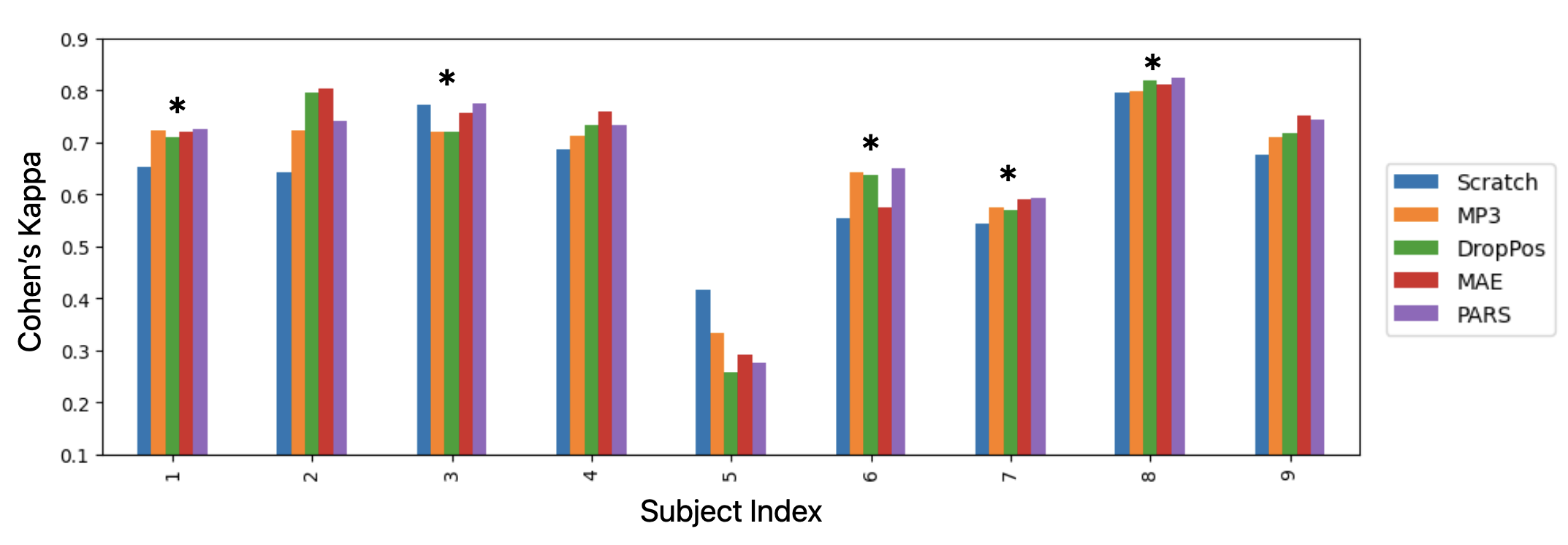}
  \caption{\textbf{Wearable Sleep Staging Results.} As described in Section \ref{section:transfer_learning}, each pretrained model is fine-tuned using a leave one subject out cross-validation approach. The per-subject Cohen's Kappa values for each pretraining strategy is reported in the bar chart above. A $*$ is placed above the five subjects where PARS outperformed the four other approaches. Note that all pretraining strategies perform poorly on Subject 5. It is reported by \cite{Mikkelsen2017} that there was a deterioration in the electrode-body contact in this subject for one ear. Better rejection of noisy channels will be investigated in future work.}
  \label{fig:eesm_results}
\end{figure}


\end{document}